\documentclass[sigconf]{acmart}
\renewcommand\footnotetextcopyrightpermission[1]{}
\AtBeginDocument{%
  }

\setcopyright{acmcopyright}
\copyrightyear{2023}
\acmYear{2023}
\acmDOI{XXXXXXX.XXXXXXX}

\acmConference[ACM Multimedia '23]{Make sure to enter the correct
  conference title from your rights confirmation emai}{October 29 – November 3, 2023}{Ottawa, Ontario, Canada}
\acmPrice{15.00}
\acmISBN{978-1-4503-XXXX-X/18/06}

\usepackage{multirow}
\usepackage{bbding}
\usepackage{colortbl}
\usepackage{color}

\acmSubmissionID{xxxx} 

\begin{document}

\title{PM-DETR: Domain Adaptive Prompt Memory for \\Object Detection with Transformers}

\author{Peidong Jia}
\authornote{Both authors contributed equally to this research.}
\author{Jiaming Liu}
\authornotemark[1]
\affiliation{%
  \institution{Peking University}
  \country{China}
}

\author{Senqiao Yang}
\affiliation{%
  \institution{Harbin Institute of Technology, Shenzhen}
  \country{China}}

\author{Jiarui Wu}
\affiliation{%
  \institution{Beihang University}
  \country{China}
}

\author{Xiaodong Xie}
\author{Shanghang Zhang}
\affiliation{%
  \institution{Peking University}
  \country{China}
}

\renewcommand{\shortauthors}{Trovato et al.}

\begin{abstract}
The Transformer-based detectors (i.e., DETR) have demonstrated impressive performance on end-to-end object detection. However, transferring DETR to different data distributions may lead to a significant performance degradation. Existing adaptation techniques focus on model-based approaches, which aim to leverage feature alignment to narrow the distribution shift between different domains. In this study, we propose a hierarchical Prompt Domain Memory (PDM) for adapting detection transformers to different distributions. PDM comprehensively leverages the prompt memory to extract domain-specific knowledge and explicitly constructs a long-term memory space for the data distribution, which represents better domain diversity compared to existing methods. 
Specifically, each prompt and its corresponding distribution value are paired in the memory space, and we inject top M distribution-similar prompts into the input and multi-level embeddings of DETR. 
Additionally, we introduce the Prompt Memory Alignment (PMA) to reduce the discrepancy between the source and target domains by fully leveraging the domain-specific knowledge extracted from the prompt domain memory. Extensive experiments demonstrate that our method outperforms state-of-the-art domain adaptive object detection methods on three benchmarks, including scene, synthetic to real, and weather adaptation. Codes will be released.
\end{abstract}

\begin{teaserfigure}
  \includegraphics[width=0.99\textwidth]{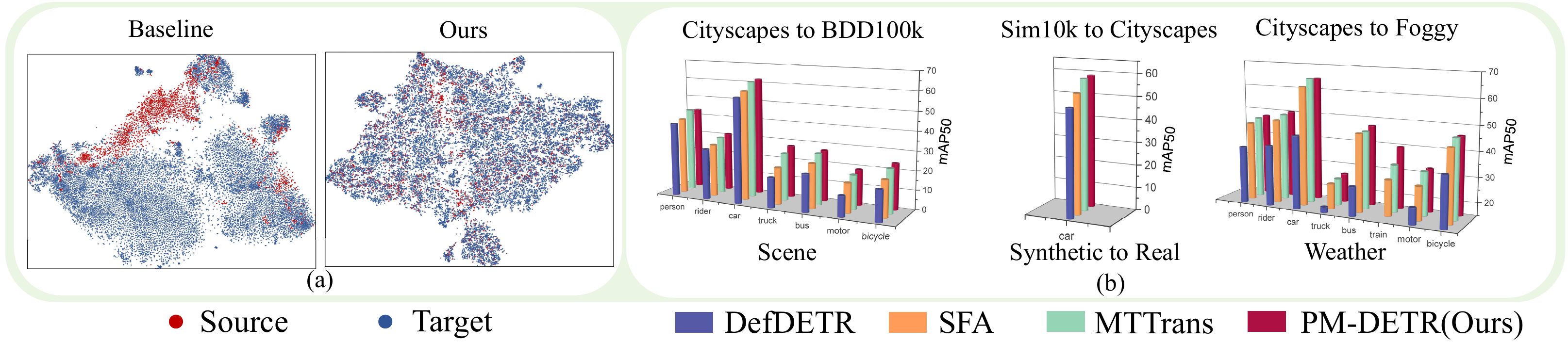}
  \vspace{-0.5 cm}
  \caption{(a) compares the t-SNE results of different methods on the source and target domain data, and our method aligns the domain shift well compared to the baseline method. (b) indicates that our method achieves state-of-the-art (SOTA) performance on three challenging domain adaptation benchmarks.}
  \label{fig:teaser}
\end{teaserfigure}


\maketitle
\section{Introduction}
Object detection is a crucial computer vision task and serves as a prerequisite for various real-world applications, such as autonomous driving \cite{arnold2019survey, chen2017multi, huang2020autonomous}, visual grounding \cite{liu2019learning, yang2021sat}, and manipulation \cite{schwarz2018rgb,zhang2021depth}. Convolutional Neural Networks (CNN) detectors \cite{ren2015faster,redmon2016you,liu2016ssd} have shown satisfactory results, but they heavily depend on hand-crafted operations like non-maximum suppression. In recent times, a series of DEtection TRansformer (DETR) methods \cite{carion2020end, zhu2020deformable} have been proposed with an end-to-end pipeline, which delivers promising performance when the test data is from the same distribution as the training data. However, such a fixed distribution is not typical in real-world scenarios \cite{Radosavovic2022}, which often comprise diverse and disparate domains. When applying pre-trained DETR models, distribution shift commonly occurs \cite{sakaridis2021acdc}, leading to significant performance degradation on the target data.

Existing adaptation techniques for DETR mainly rely on model-based approaches \cite{wang2021exploring, yu2022mttrans}, which aim to narrow the distribution shift between different domains via sequence feature alignment. Recent developments in prompt learning for both natural language processing (NLP) \cite{lester2021power, li2021prefix, liu2023pre} and computer vision \cite{jia2022visual} have motivated researchers to introduce visual prompts in domain adaptation tasks. Several recent studies \cite{gan2022decorate, chen2022multi, gao2022visual, yang2023exploring} have leveraged prompts randomly set at the image or feature-level and fine-tuned them to extract domain-specific or maintain domain-invariant knowledge. These approaches offer a prompt-based perspective to address distribution shift, which can further aid the model-based methods in achieving better representations in the target domain.
However, when prompt-based methods are applied in the target domain with various scene conversion and complex distribution data (i.e., autonomous driving data), the prompts are difficult to learn the long-term domain knowledge for the full data. 
Meanwhile, since object detection is an instance-level task and exits multiple objects in each sample, the previous prompt methods are hard to extract diverse domain knowledge for each category.

To this end, we propose a hierarchical Prompt Domain Memory (PDM) for adapting detection transformers, which can extract domain-specific knowledge by learning a set of prompts that dynamically instruct the DETR. Specifically, each prompt and its corresponding distribution value are paired in the memory space, and we dynamically select top M distribution similar prompts for each sample using the value. PDM explicitly constructs a long-term memory space for the detection transformer, allowing DETR to learn complex data distribution and different category domain knowledge at multi-levels, including input, token, and query levels.
With the help of PDM, as shown in Fig.\ref{fig:teaser} (a), feature representations in the two domains achieve smaller distribution shift compared to the previous method. However, while the prompt memory can extract more comprehensive domain-specific knowledge, it cannot reduce the distribution distance between different domains \cite{gan2022decorate}. To address this limitation, we propose the Prompt Memory Alignment (PMA) method, which reduces the discrepancy between source and target domains in the Unsupervised Domain Adaptation (UDA) task.
Traditional feature alignment methods \cite{wang2021exploring, yu2022mttrans} can only align a small number of different domain samples in each iteration due to the limited GPU memory. Different from previous methods, since the proposed prompt memory can better represent the diversity of each domain, the PMA can fully leverage the domain-specific knowledge extracted from the memory and efficiently address the distribution shift. In addition, along with introducing PDM, we make the first attempt to design the visual prompt alignment strategy to jointly address the domain shift problem.
In conclusion, our proposed approach of PDM and PMA enhances the performance of detection transformers in adapting to target domains by extracting diverse domain-specific knowledge and reducing the discrepancy between source and target domains.

We evaluate the prompt-based PM-DETR on three challenging benchmarks of UDA, including scene adaptation (Cityscapes \cite{cordts2016cityscapes} to BDD100k \cite{yu2018bdd100k}), synthetic to real adaptation (Sim10k \cite{johnson2016driving} to Cityscapes), and weather adaptation (Cityscapes to Foggy Cityscapes \cite{sakaridis2018semantic}). Our method outperforms state-of-the-art (SOTA) domain adaptive object detection methods, which improves the result to 58.6\%, 33.3\%, and 44.3\% mAP in the three benchmarks, shown in Fig.\ref{fig:teaser} (b). 

The main contributions are summarized as follows:

\textbf{1)} We propose a hierarchical Prompt Domain Memory (PDM) to adapt detection transformers to different distributions, which constructs a long-term memory space to fully learn the complex data distribution and diversity domain-specific knowledge.

\textbf{2)} In order to better apply PDM in the Unsupervised Domain Adaptation (UDA), we propose the Prompt Memory Alignment (PMA) method to reduce the distribution distance between two domains, which can fully leverage the domain-specific knowledge extracted from the memory space.

\textbf{3)} We conduct extensive experiments on three challenging UDA scenarios to evaluate the effectiveness of our method. The method achieves SOTA performance in all scenarios, including scene, synthetic to real, and weather adaptation. 


\section{Related Works}

\subsection{Object Detection}

Object detection is a critical task of computer vision~\cite{zheng2015scalable, he2017mask, kirillov2019panoptic}. Previous convolutional neural network (CNN)-based approaches can be broadly categorized into two groups: the more complex two-stage methods~\cite{ren2015faster, lin2017feature, yang2019reppoints} and the lighter one-stage methods~\cite{redmon2016you, liu2016ssd, tian2019fcos}. However, these approaches exhibit a significant limitation due to their heavy reliance on handcrafted processes and initial guesses, particularly the non-maximum suppression (NMS) post-processing, which hinders their ability to be trained end-to-end. Recent advancements, such as DETR~\cite{carion2020end} and Deformable DETR~\cite{zhu2020deformable}, have addressed this issue by incorporating vision Transformers~\cite{vaswani2017attention}.
Deformable DETR introduces an innovative deformable multi-head attention mechanism that enables sparsity in attention and multi-scale feature aggregation without necessitating a feature pyramid structure. This innovation results in faster training and enhanced performance. For the critical issue of DETR, slow training convergence, Conditional DETR \cite{meng2021conditional} speed up DETR training by leveraging a conditional cross-attention mechanism. In order to more effectively utilize the attention mechanism, SMCA~\cite{gao2021fast} proposes a co-attention scheme that expedites DETR convergence, which includes multi-head and scale-selection attention.
Moreover, DN-DETR~\cite{li2022dn} offers a novel perspective on faster training by employing a denoising approach to improve the stability of bipartite graph matching during the training stage. In our study, we use the classical Deformable DETR as the base detector and make the first attempt to introduce domain prompts into its workflow. We also design a hierarchical domain prompt memory to facilitate diverse domain knowledge extraction.

\subsection{Domain adaptive object detection}

Domain Adaptive Faster R-CNN~\cite{chen2018domain} established the foundation for investigating domain-adaptive object detection techniques. Subsequent studies primarily utilized the adversarial training paradigm for cross-domain feature alignment. Various strategies have been proposed in these approaches to aggregate image or instance features, such as leveraging categorical predictions~\cite{xu2020exploring,xu2020cross} and exploiting spatial correlations~\cite{xu2020cross,cai2019exploring}. Hierarchical alignment of features was conducted at multiple levels, encompassing global, local, instance, and category levels~\cite{xu2020exploring,saito2019strong,xu2020cross,cai2019exploring,luo2021unsupervised}. Recent advancements in this field introduced innovative methods, including PICA~\cite{zhong2022pica}, which specializes in few-shot domain adaptation, and Visually Similar Group Alignment (ViSGA)\cite{rezaeianaran2021seeking}, which employs similarity-based hierarchical agglomerative clustering, achieving exceptional performance on specific benchmarks. Moreover, several studies explored alternative domain adaptation techniques or utilized different base detectors, such as Mean Teacher with Object Relations (MTOR)\cite{cai2019exploring} and Unbiased Mean Teacher (UMT)~\cite{deng2021unbiased}. Regarding the Transformer object detector, existing adaptation techniques for DETR predominantly rely on model-based approaches~\cite{wang2021exploring, yu2022mttrans}, aiming to reduce the distribution shift between different domains through sequence feature alignment. In this paper, we provide a novel perspective on DETR cross-domain transfer by introducing a prompt-based approach. Specifically, we propose a Prompt-based Domain Memory (PDM) and Prompt Pool Alignment (PMA) to enhance the performance of detection transformers in adapting to target domains. This is achieved by extracting diverse domain-specific knowledge and minimizing the discrepancy between source and target domains.

\subsection{Prompt Learning} 

Prompt learning, originally introduced in the field of natural language processing (NLP), aims to adapt pre-trained language models to various downstream tasks in a parameter-efficient manner~\cite{liu2023pre, Brownetal2020, Radfordetal2021, Yaoetal2021, zhouetal2022cocoop, zhouetal2022coop}. Recently, researchers have extended the paradigm of prompt learning to efficient fine-tuning of vision models~\cite{Bahngetal2022}. VPT~\cite{jia2022visual} and its variants~\cite{Conderetal2022, Sandleretal2022} introduce minimal trainable parameters at the image or feature level of Transformer-based models for efficient transfer learning. L2P~\cite{ZifengWangetal2021} and its follow-up method~\cite{wang2022dualprompt} propose a prompt pool-based approach for continual learning, aiming to avoid catastrophic forgetting and error accumulation. 
More recently, visual prompts have shown promising results in domain adaptation. DAPL~\cite{Geetal2022} made the initial attempt to incorporate visual prompts into unsupervised domain adaptation (UDA). Subsequent studies, such as~\cite{chen2022multi, gao2022visual, gan2022decorate}, explored diverse approaches to leverage visual prompts for classification domain adaptation problems. Additionally, SVDP~\cite{yang2023exploring} proposed a sparse visual prompt for efficient adaptation in segmentation tasks. However, these studies primarily focus on image-level~\cite{gan2022decorate} and pixel-level~\cite{yang2023exploring} domain adaptation tasks and are not optimized for instance-level transformer detection. Furthermore, when prompt-based methods are applied in the target domain with various scene conversions and complex data distributions~\cite{yu2022mttrans} (e.g., autonomous driving data), it becomes challenging to utilize previous lightweight methods~\cite{gan2022decorate} to learn long-term domain knowledge. Due to the instance-level property, with multiple objects present in each sample, previous prompt methods~\cite{gan2022cloud} struggle to extract diverse domain knowledge for each category. To address this issue, we design a Prompt-based Domain Memory (PDM) tailored for adapting detection transformers, particularly in scenarios involving diverse data distributions.

\section{Methods}
\noindent\textbf{Preliminary} This section introduces PM-DETR for transformer-based domain adaptive detectors. Given a model $D_S(y|x)$ trained in labeled source domain samples $\mathcal{D}_S(x,y)$, where x and y represent input data and Ground Truth, respectively. Our goal is adapting $D_S$ to target model $D_T$ through unlabeled target domain data $\mathcal{D}_T(x)$. 

\begin{figure*}[htbp]
  \centering
  \includegraphics[width=0.95\linewidth]{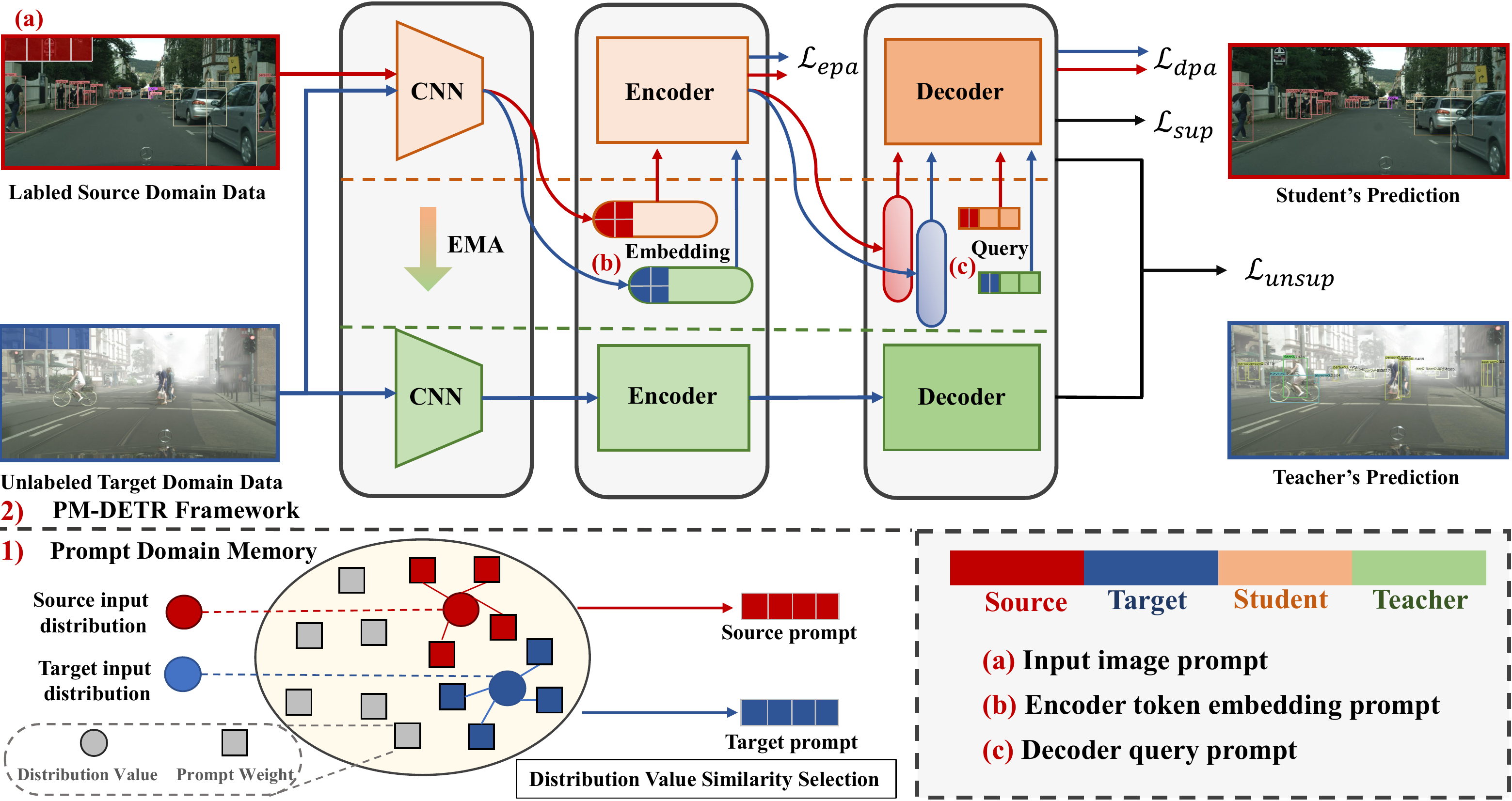}
\vspace{-0.3 cm}  
  \caption{The Overall framework of PM-DETR. \textcolor{red}{(1)} \textbf{Prompt Domain Memory (PDM).} Hierarchical prompt domain memory, which warps on input, encoder token, and decoder query, learns a set of prompts for excavating diverse domain-specific knowledge. We adaptively select the top M prompts by distribution value similarity. The selection strategy as described in Eq. \ref{eq:promptsel}.   \textcolor{red}{(2)} \textbf{PM-DETR Framework.} We construct a teacher-student paradigm to optimize the student model and PDM by source domain labels and target domain pseudo labels. Besides, we propose a novel prompt memory alignment (PMA) to constraint multi-level prompts digging in-domain knowledge and spatial correlations, as described in Eq. \ref{eq: PAL}.  }
  \label{fig:pipeline}
  \vspace{-0.2 cm} 
\end{figure*}

We propose a comprehensive prompt-based method to enhance the performance of detection transformers in adapting to target domains by extracting diverse domain-specific knowledge and reducing the discrepancy between source and target domains.
In section \ref{sec:moti}, we first verify the motivation of the prompt-based adaptation method for transformer detectors.
In section \ref{sec:HPM}, a hierarchical Prompt Domain Memory (PDM) is illustrated to extract domain-specific knowledge in the multi-level transformer latent space. In section \ref{sec:PMA} , Prompt Memory Alignment (PMA) is proposed to reduce the discrepancy between source and target domains. Finally, in Section \ref{sec:OOP} , we elaborate on our training policy in detail. The overall pipeline is shown in Fig. \ref{fig:pipeline}, and Deformable DETR \cite{zhu2020deformable} is utilized as our default detection network.

\subsection{Motivation of Prompt-based Method}
\label{sec:moti}

\begin{figure}[t]
  \centering
  \includegraphics[width=0.85\linewidth]{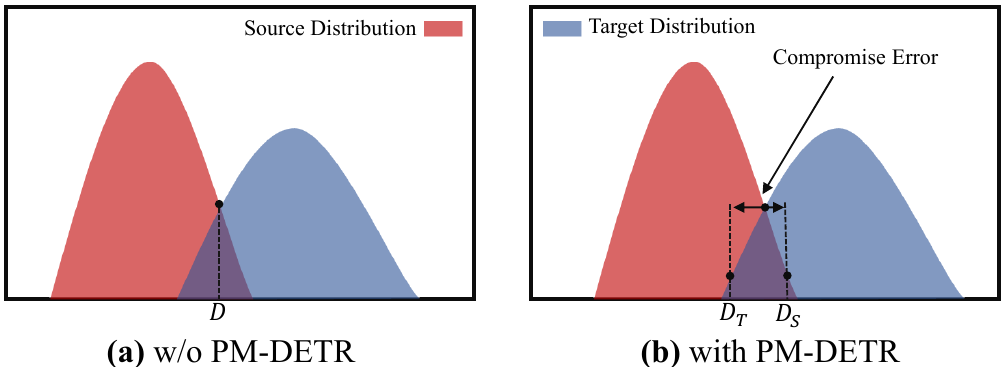}
    \vspace{-0.32 cm}
  \caption{Demonstrate the compromise error. The red and blue areas represent the probability of correct classification in the source and target domains. When the models of the two domains are identical, model will converge to $D$. Our decoupling method converges to $D_S$ and $D_T$ respectively, avoiding the compromise error.}
  \vspace{-0.35 cm}
  \label{fig:motivation1}
  
\end{figure}
We verify the motivates of introducing a prompt-based method and constructing a long-term prompt domain memory. First, there is a brief explanation of how the traditional model-based adaptation method tackles the Unsupervised Domain Adaptation (UDA) problem.
These methods \cite{ganin2016domain, hu2022prosfda, yu2022mttrans} pursue to shrink upper boundary of the target error $err_T$
by the sum of the source error $err_S$ and a notion of distance $d_\mathcal{H}$ between the source and the target distributions in hypothesis space $\mathcal{H}$. Since $err_S$ is primarily influenced by model complexity, researchers have predominantly focused on minimizing the inter-domain distance $d_\mathcal{H}$. 
Suppose that construct a unified dataset $\mathbb{U}$ defined as below:

\begin{equation}
  \mathbb{U} = \{x_i, y=0\}_{i=1}^{p} \cup \{x_j, y=1\}_{j=p+1}^{q}
\end{equation}

Where the first $p$ samples are from source domain $\mathcal{D}_S$ and labeled as 0, the rest samples are from target domain $\mathcal{D}_T$ and labeled as 1. 
By constructing this unified dataset $\mathbb{U}$, we create a share high dimensional space for both the source and target domains, allowing us to effectively measure the distance between their distributions in the hypothesis space $\mathcal{H}$, shown in Fig. \ref{fig:motivation1} (a). This unified dataset serves as a foundation for minimizing the inter-domain distance and enabling better adaptation from the source to the target domain. Furthermore, the work of Ben-David et al. \cite{ben2006analysis, ben2010theory} has provided evidence that the empirical $\mathcal{H}$-divergence between two domains can be computed by the following equation:

\begin{equation}
\begin{aligned}
  d_{\mathcal{H}}(S,T) &= 2 (1 -  \min_{D \in \mathcal{H}} \lbrack \frac{1}{p}\sum_{i=1}^{p} D[\mathbb{U}(y_i=0)] \\
  &+ \frac{1}{q-p}\sum_{j=p+1}^{q-p} D[\mathbb{U}(y_j=1)] \rbrack) \\
  &= 2 (1 - \varepsilon_{D}^{S} - \varepsilon_{D}^{T})
\end{aligned}
\end{equation}

The methods discussed above are based on an intuitive assumption that model parameters capable of fitting both the source and target domain data well can be learned in the same hypothesis space $\mathcal{H}$. However, it is important to note that this assumption does not always hold true in practical scenarios. In reality, there can be inherent differences between the source and target domains that make it challenging to find a single hypothesis space that adequately captures both domains. As a result, when attempting to adapt a model from the source to the target domain, a compromise error may be introduced. Fig. \ref{fig:motivation1} visually illustrates this compromise error, which represents the discrepancy between the optimal models for each domain and the compromise model that attempts to accommodate both domains. 

\begin{figure}[t]
  \centering
  \includegraphics[width=\linewidth]{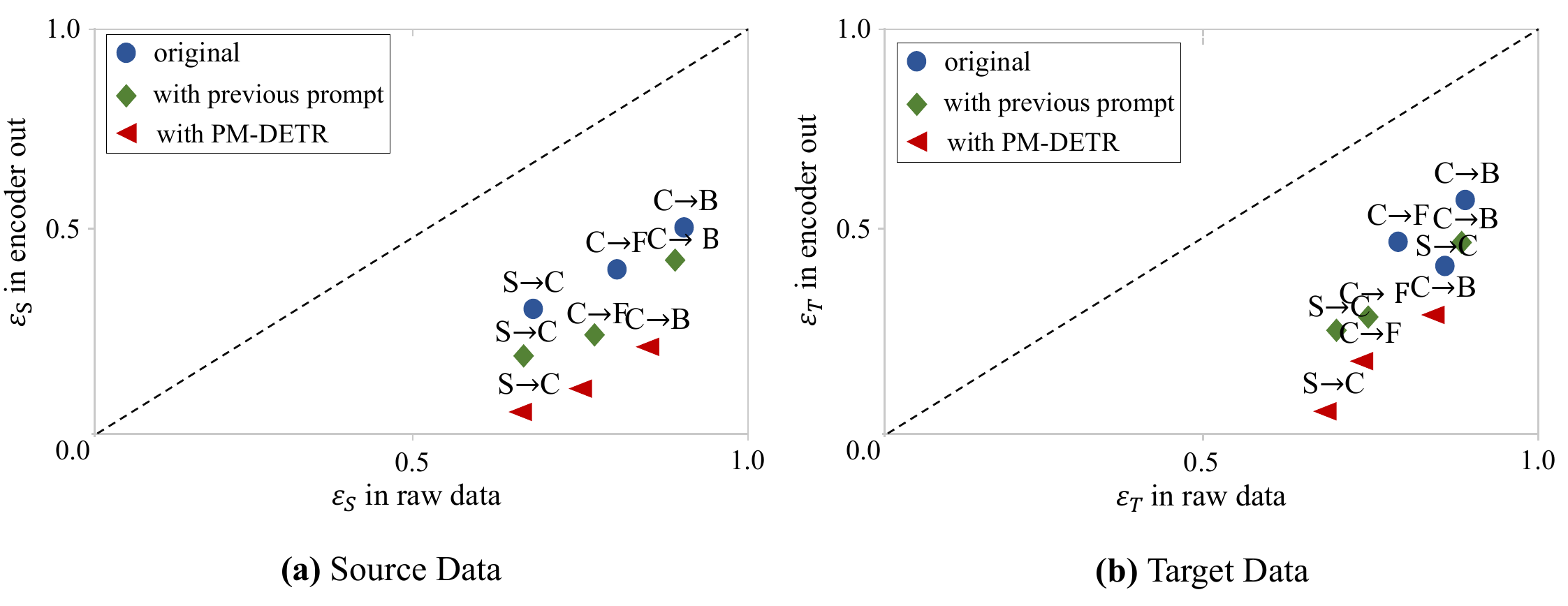}
  \vspace{-0.82 cm}
   \caption{Quantitative domain classification errors comparison of baseline and our method. Where $C\xrightarrow{}B$ means from Cityscapes to BDD100k, $S\xrightarrow{}C$ means from Sim10k to Cityscapes, $C\xrightarrow{}F$ means from Cityscapes to Foggy.}
  \label{fig:motivation2}
  \vspace{-0.46 cm}
\end{figure}


To this end, drawing inspiration from soft prompt learning techniques used in NLP and computer vision tasks, where the pre-trained model is adapted to different downstream tasks through prompt manipulation within the input sequence, we propose that incorporating a lightweight visual prompt can assist in bounding the inter-domain distance $d_{\mathcal{H}}$ within a smaller interval. Specifically, we adopt domain prompt warp into input image, encoder embedding, and decoder queries, so that the hypothesis space in source and target domain can be decoupled to $\mathcal{H}_S$ and $\mathcal{H}_T$. The prompt memory explicitly constructs a long-term memory space for better representing the diversity of domain knowledge, which further assist the hypothesis space decoupling. Under different hypothesis spaces, prompt memory alignment encourages mining in-domain knowledge by constraints, so the model is optimized as ${D_S}$ and ${D_T}$ in the source and target domains, respectively, as shown in Fig. \ref{fig:motivation1} (b). Thus, in theory, the inter-domain distance will be reduced by the following equation

\begin{equation}
\begin{aligned}
  d_{{\mathcal{H}_S, \mathcal{H}_T}}(S,T) &= 2 (1 - \varepsilon_{D_S}^{S} - \varepsilon_{D_T}^{T}) <  d_{\mathcal{H}}(S,T)
\end{aligned}
\end{equation}

To further substantiate the presence of compromise error, we conduct experiments on three domain adaptive object detection tasks as depicted in Fig. \ref{fig:motivation2}. The figure quantitatively compares the classification errors of the model-based, previous prompt-based, and Prompt Domain Memory (PDM) method on a domain classifier consisting of three multi-layer perceptrons in series. As we can see, the previous prompt-based method achieve smaller classification errors compared to the model-based method, and the proposed PDM further has a significant improvement in classification errors. As proven by \cite{ganin2016domain}, generalization upper boundary on the target risk can be smaller attributed to lower classification error, which means that model will perform better in the target domain. 

\subsection{Hierarchical Prompt Domain Memory}
\label{sec:HPM}

We propose prompt domain memory $\mathbf{P}_S$ and $\mathbf{P}_T$ 
for source and target domains, respectively. Each memory pool $\mathbf{P}$
has $N$ prompt pairs $\{<\mathbf{v_i},\mathbf{p_i}>|i=1,...,N,\mathbf{v} \in \mathbb{R}^{1 \times d}, \mathbf{p} \in \mathbb{R}^{L \times d}\}$, where $\mathbf{v}$ 
indicates prompt distribution value and $\mathbf{p}$ indicates visual prompt weight. $L$ stands for embedding length and $d$ stands for embedding dimention.
We randomly initialize $\mathbf{v}$ and $\mathbf{p}$. Visual prompt weight $\mathbf{p}$ 
stands for the aggregation of domain-specific knowledge, which can be learned by warping in the prefix of the input sequence. 
Prompt distribution value $\mathbf{v}$ will be used to measure the distribution similarity with the input. In this way, prompt memory can cover diverse domain knowledge for complex data distribution. 


\textbf{Hierarchical Warp Position. } We introduce prompt memory 
in three crucial embeddings including 
input image, encoder token, and decoder queries. The motivations are three-fold. 
First, the combination of multi-level prompts provides a comprehensive domain transfer mechanism, where different levels of prompts decrease domain shift that could not be narrowed by previous levels.
Second, prompt in decoder query can align the distribution of objects in different domain datasets, 
thus improving the recall of Deformable DETR. Third, the multi-level prompts only increase the number of parameters by a very small amount (0.063\% of model parameters), but it can greatly enhance the plasticity of the model and release the power in learning diverse domain-specific representations. 
Our ablation experiments in Sec. 4.2 show that the hierarchical prompt memory can better represent the domain diversity and boost the performance of Deformable DETR in the target domain. 


\textbf{Distribution Value Similarity Selection. } The image scenes, object classes, 
and object distributions in the target dataset have large variances, 
which often leads to sub-optimal performance if only use the 
same prompt for each instance to extract domain knowledge. 
We design a distribution-guided strategy to adaptively 
select prompts from prompt memory. We project input embedding 
by transformation function $\gamma$ to $v$'s shape. Here we utilize 
the average mean along the embedding channel to aggregate input characteristics. 
Then we calculate the cosine similarity between $v$ and the projection embedding utilizing function $\psi$.

\begin{equation}
\label{eq:promptsel}
    \mathbf{V_M} = \mathrm{argmax} \sum_{i=1}^{M} \psi (\mathbf{V}, \gamma(x))
\end{equation}


According to the cosine similarity value, $\mathbf{V_M}$ is the selected nearest M neighbor prompts (i.e. $M=4$) in prompt memory.

\subsection{Prompt Memory Alignment}
\label{sec:PMA}


After receiving the domain knowledge transferred
from prompt memory, we further introduce Prompt Memory 
Alignment (PMA) to address the domain shift accumulation. 
For the encoder phase, we aim to pull close the input and token level prompts from two domain prompt memory, as shown in Fig .\ref{fig:pipeline}.
Specifically, we utilize respective MLPs to project prompt 
tokens to a shared embedding space, in which the dimension 
is $\mathbf{L}\times C\times 2$, $\mathbf{L}$ equals to encoder token length, channel dimension $C$ is set to 256.  We
adopt encoder prompt alignment loss $\mathcal{L}_{epa}$ to pull close the two domain prompt embeddings and explicitly constrain prompt to learn in-domain knowledge, as shown in Eq. \ref{eq: PAL}.
\begin{equation}
\label{eq: PAL}
\mathcal{L}_{epa}(X, D) = \lambda_1 \min D(\mathbf{X}_{i<|\mathbf{p}|\times M}) + \lambda_2 \max D(\mathbf{X}_{i \geq |\mathbf{p}|\times M})
\end{equation}

Where $D$ denotes the domain discriminator. For decoder phase, decoder prompt alignment loss $\mathcal{L}_{dpa}$, similar to $\mathcal{L}_{epa}$, are proposed. Since objects in the same category and spatially connected tend to be visually similar, $\mathcal{L}_{dpa}$ constraint prompts in decoder queries to learn from categorical and spatial correlations while decreasing data distribution distance between the two domains. 



\subsection{Overall Optimization for PM-DETR}
\label{sec:OOP}



PM-DETR leverages the teacher-student framework, which includes two models with the same architecture and weights at initialization During training, the student model is updated using back-propagation, while the teacher model is updated by taking the Exponential Moving Average (EMA) of the student's weights. The weights of the teacher model $\theta_{t}^{\prime}$ at time step $t$ is calculated by taking a weighted average of the teacher's previous weights and the current student's weights $\theta_t$:

\begin{equation}
 \theta_{t}^{\prime} = \alpha\theta_{t-1}^{\prime} + (1-\alpha)\theta_{t}
\label{eq: ema}
\end{equation}

$\alpha$ is a smoothing coefficient hyperparameter and is set to 0.999. And the object query embeddings are kept the same between the two models to enhance consistency. The object queries are trainable embeddings, initialized with the normal distribution at the start of the training procedure. Then we use the temporally ensembled teacher model to optimize the parameter of the student model and prompts in the target domain via pseudo labels. 




For the integral optimizing, the first loss is a penalty on the source domain for supervised learning ($\mathcal{L}_{sup}$), which distills domain independent generic features and avoids catastrophic forgetting. The second loss uses the target domain pseudo-labels generated by the teacher model for unsupervised learning ($\mathcal{L}_{unsup}$) to extract the target domain knowledge. The two losses are separately used to optimize source and target domain prompt memory. Combined with the proposed prompt alignment losses, the overall constraint function is:

\begin{equation}
    \mathcal{L} = \lambda_{s}\mathcal{L}_{sup} + \lambda_{us}\mathcal{L}_{unsup} + \lambda_{epa}\mathcal{L}_{epa} + \lambda_{dpa}\mathcal{L}_{dpa}
\end{equation}
To maintain the balance of loss penalties, $\lambda_{s}$ and $\lambda_{us}$ are set to 1, $\lambda_{epa}$, and $\lambda_{dpa}$ are set to 0.25. The detection loss ($\mathcal{L}_{sup}$ and $\mathcal{L}_{unsup}$) are combined by focal loss and $\mathrm{L}1$ loss \cite{carion2020end, zhu2020deformable}. 

\definecolor{gbypink}{rgb}{0.99, 0.91, 0.95} 
\begin{table*}[t]
\scriptsize
\begin{center}
\caption{Performance comparison of different methods for weather adaptation, that is, from Cityscapes to Foggy Cityscapes. FRCNN and DefDETR are abbreviations for Faster R-CNN and Deformable DETR, respectively.}
\vspace{-0.3 cm}
\setlength{\tabcolsep}{3.2mm}{
\begin{tabular}{c|c|c|cccccccc|c|c}
\hline
Method                 & Detector & Publication & person & rider & car  & truck & bus  & train & mcycle & bicycle & mAP & Gain \\\hline\hline
\multicolumn{13}{l}{\textit{Two Stage :}}\\\hline
FasterRCNN~\cite{ren2015faster}(Source)     & FRCNN  & NIPS2015 & 26.9   & 38.2  & 35.6 & 18.3  & 32.4 & 9.6   & 25.8   & 28.6    & 26.9 & 00.00\\
CR-DA~\cite{xu2020exploring}         & FRCNN & CVPR2020  & 30.0     & 41.2  & 46.1 & 22.5  & 43.2 & 27.9  & 27.8   & 34.7    & 34.2 & +07.3\\
DivMatch~\cite{kim2019diversify}      & FRCNN  & CVPR2019  & 31.8   & 40.5  & 51.0   & 20.9  & 41.8 & 34.3  & 26.6   & 32.4    & 34.9 & +08.0 \\
MTOR~\cite{cai2019exploring}       & FRCNN  & CVPR2019 & 30.6   & 41.4  & 44.0   & 21.9  & 38.6 & 40.6  & 28.3   & 35.6    & 35.1 & +08.2\\
SWDA~\cite{saito2019strong}           & FRCNN  & CVPR2019  & 31.8   & 44.3  & 48.9 & 21.0    & 43.8 & 28    & 28.9   & 35.8    & 35.3 & +08.4\\
SCDA~\cite{zhu2019adapting}          & FRCNN  & CVPR2019  & 33.8   & 42.1  & 52.1 & 26.8  & 42.5 & 26.5  & 29.2   & 34.5    & 35.9 & +09.0\\
CR-SW~\cite{xu2020exploring}       & FRCNN  & CVPR2020 & 34.1   & 44.3  & 53.5 & 24.4  & 44.8 & 38.1  & 26.8   & 34.9    & 37.6 & +10.7\\
GPA~\cite{xu2020cross}          & FRCNN  & CVPR2020 & 32.9   & 46.7  & 54.1 & 24.7  & 45.7 & 41.1  & 32.4   & 38.7    & 39.5 & +12.6\\
ViSGA~\cite{rezaeianaran2021seeking}  & FRCNN & ICCV2021  & 38.8   & 45.9  & 57.2 &  29.9  & 50.2 & 51.9  & 31.9   & 40.9    & 43.3 & +16.4\\\hline
\multicolumn{13}{l}{\textit{One Stage :}}\\\hline
FCOS~\cite{tian2019fcos} (Source)           & FCOS  & ICCV2019    & 36.9   & 36.3  & 44.1 & 18.6  & 29.3 & 8.4   & 20.3   & 31.9    & 28.2 & +01.3\\
EPM\cite{hsu2020every}         & FCOS   & ECCV2020  & 44.2   & 46.6  & 58.5 & 24.8  & 45.2 & 29.1  & 28.6   & 34.6    & 39.0  & +12.1 \\\hline
\multicolumn{13}{l}{\textit{Transformer based :}}\\\hline
Def DETR~\cite{zhu2020deformable} (Source) & DefDETR & ICLR2021 & 37.7   & 39.1  & 44.2 & 17.2  & 26.8 & 5.8   & 21.6   & 35.5    & 28.5 & +01.6\\
SFA~\cite{wang2021exploring}        & DefDETR & ACMMM2021 & 46.5   & 48.6  & 62.6 & 25.1  & 46.2 & 29.4  & 28.3   & 44.0      & 41.3 & +14.4\\
MTTrans\cite{yu2022mttrans}              & DefDETR & ECCV2022 &  47.7   &  49.9  &  65.2 & 25.8  & 45.9 & 33.8  &  32.6  & 46.5     & 43.4 & +16.5\\
\rowcolor{lightgray} PM-DETR(Ours)    & DefDETR & - &  \textcolor{black}{\textbf{47.8}}   & \textcolor{black}{\textbf{50.2}}  & \textcolor{black}{\textbf{64.7}} & \textcolor{black}{\textbf{26.5}}  & \textcolor{black}{\textbf{47.2}} & \textbf{39.6}  &  \textcolor{black}{\textbf{32.4}}  & \textcolor{black}{\textbf{46.1}}      &\cellcolor{gbypink}\textcolor{red}{\textbf{44.3}} &\cellcolor{gbypink}\textcolor{red}{\textbf{+17.4}}\\
\hline
\end{tabular}}
\label{tab:1}
\end{center}
\vspace{-8pt}
\end{table*}

\begin{table*}[t]
\scriptsize
\begin{center}
\caption{Performance comparison of different methods for the scene adaptation, i.e., Cityscapes to BDD100k daytime subset.}
\vspace{-0.3 cm}
\setlength{\tabcolsep}{3.6mm}{
\begin{tabular}{c|c|c|ccccccc|c|c}
\hline
Methods                & Detector &  Publication & person & rider & car  & truck & bus  & mcycle & bicycle & mAP & Gain \\\hline\hline
\multicolumn{12}{l}{\textit{Two Stage :}}\\\hline
FasterR-CNN~\cite{ren2015faster}(Source)    & FRCNN   & NIPS2015 & 28.8   & 25.4  & 44.1 & 17.9  & 16.1 & 13.9   & 22.4    & 24.1 & 0.00 \\
DAF~\cite{chen2018domain}           & FRCNN & CVPR2018 & 28.9   & 27.4  & 44.2 & 19.1  & 18.0   & 14.2   & 22.4    & 24.9 & +0.8\\
SCDA~\cite{zhu2019adapting}         & FRCNN  & CVPR2019 & 29.3   & 29.2  & 44.4 & 20.3  & 19.6 & 14.8   & 23.2    & 25.8 & +1.7\\
CR-DA~\cite{xu2020exploring}     & FRCNN  & CVPR2020 & 30.8   & 29.0    & 44.8 & 20.5  & 19.8 & 14.1   & 22.8    & 26.0  & +1.9 \\
SWDA~\cite{saito2019strong}        & FRCNN   & CVPR2019 & 29.5   & 29.9  & 44.8 & 20.2  & 20.7 & 15.2   & 23.1    & 26.2 & +2.1\\
CR-SW~\cite{xu2020exploring}   & FRCNN  & CVPR2020 & 32.8   & 29.3  & 45.8 & 22.7  & 20.6 & 14.9   & \textbf{25.5}    & 27.4 & +3.3\\\hline
\multicolumn{12}{l}{\textit{One Stage :}}\\\hline
FCOS~\cite{tian2019fcos}(Source)   & FCOS  & ICCV2019  & 38.6   & 24.8  & 54.5 & 17.2  & 16.3 & 15.0     & 18.3    & 26.4 & +2.3\\
EPM~\cite{hsu2020every}      & FCOS   & ECCV2020 & 39.6   & 26.8  & 55.8 & 18.8  & 19.1 & 14.5   & 20.1    & 27.8 & +3.7\\\hline
\multicolumn{12}{l}{\textit{Transformer Based :}}\\\hline
Def DETR~\cite{zhu2020deformable}(Source) & DefDETR & ICLR2021 & 38.9   & 26.7  & 55.2 & 15.7  & 19.7 & 10.8   & 16.2    & 26.2 & +2.1\\
SFA~\cite{wang2021exploring}  & DefDETR & ACMMM2021 & 40.2   & 27.6  & 57.5 & 19.1  & 23.4 & 15.4   & 19.2    & 28.9 & +4.8\\
MTTrans~\cite{yu2022mttrans}   & DefDETR & ECCV2022 & 44.1   & 30.1  & 61.5 & 25.1  & 26.9 & 17.7   & 23.0    & 32.6 & +8.5\\
\rowcolor{lightgray} Ours(PM-DETR)  & DefDETR & - & \textbf{43.3}   & \textbf{30.9}  & \textbf{62.0} & \textbf{27.4}  & \textbf{26.8} & \textbf{18.7}   & \textbf{23.9}    & \cellcolor{gbypink}\textcolor{red}{\textbf{33.3}} & \cellcolor{gbypink}\textcolor{red}{\textbf{+9.2}}\\

\hline
\end{tabular}}
\label{tab:3}
\end{center}
\vspace{-7pt}
\end{table*}

\section{Experiments}
We conduct extensive experiments to demonstrate the advantages of our proposed method for object detection Unsupervised Domain Adaptation (UDA) task. In Section \ref{sec: exp1}, we provide a description of the datasets, as well as the details of model settings. In Section \ref{sec: exp2}, we compare mean Average Precision (mAP) metric of PM-DETR with other baselines \cite{yu2022mttrans, zhu2020deformable, hsu2020every, tian2019fcos, ren2015faster, kim2019diversify, cai2019exploring, saito2019strong, zhu2019adapting, wang2021exploring, xu2020cross, rezaeianaran2021seeking} in three challenging domain adaptation scenarios, including Weather, Scene, and Synthetic to Real Adaptation. Comprehensive ablation studies are conducted to investigate the impact of each component in Section \ref{sec: exp3}. Furthermore in Section \ref{sec: exp4}, qualitative analysis is given to facilitate intuitive understanding. 

\subsection{Experimental Setup}
\label{sec: exp1}
\textbf{Datasets.} We evaluate our method on four public datasets, including Cityscapes \cite{cordts2016cityscapes}, Foggy Cityscapes \cite{sakaridis2018semantic}, Sim10k \cite{zheng2020cross}, and BDD100k \cite{yu2018bdd100k}. These datasets provide diverse and challenging scenarios for domain adaptation tasks:

\textbf{Weather Adaptation.} In this scenario, we use Cityscapes as the source dataset, consisting of 2,975 training images and 500 evaluation images. The target dataset is Foggy Cityscapes, generated from Cityscapes using a fog synthesis algorithm. Foggy Cityscapes introduces foggy conditions to the images, enabling us to evaluate the performance of our method in adapting object detection models from clear weather to foggy weather scenarios. 

\textbf{Scene Adaptation.} In this condition, Cityscapes serves as the source dataset, while the target dataset is the daytime subset of BDD100k. BDD100k consists of 36,728 training images and 5,258 validation images, all annotated with bounding boxes. This subset provides a diverse range of scenes captured during the daytime

\textbf{Synthetic to Real Adaptation.} In this particular scenario, we employ Sim10k as the source domain, which is generated using the Grand Theft Auto game engine. Sim10k comprises 10,000 training images, accompanied by 58,701 bounding box annotations. As for the target domain, we utilize the car instances from Cityscapes for both training and evaluation purposes.  

\noindent\textbf{Implementation Details.} Our method is built based on Deformable DETR \cite{zhu2020deformable}. We set ImageNet \cite{deng2009imagenet} pre-trained ResNet-50 \cite{he2016deep} as CNN backbone in all experiments. In the burn-in step, we adopt Adam optimizer \cite{kingma2014adam} for training over 50 epochs. The initial learning rate is set to $2e-04$, which decayed by 0.1 after 40 epochs. The batch size is set to 4 for all adaptation scenarios. In the second cross-domain training step, the model is trained for 12 epochs. The prompt-based parameters are initialized with random float numbers and set the learning rate to $2e-05$. The learning rate for the other parameters, excluding the prompt-based parameters, is relatively small and set to $2e-06$. All learning rates decayed by 0.1 after 10 epochs. In addition, we adopt mean Average Precision (mAP) with a threshold of 0.5 as the evaluation metric. We set the filtering threshold (confidence) for the pseudo-label generation to 0.5. All experiments are conducted on two NVIDIA Tesla A100 GPUs.

\subsection{Comparisons with SOTA Methods}
\label{sec: exp2}

\textbf{Weather Adaptation.} To assess the reliability of object detectors under varying weather conditions, we transfer models from Cityscapes to Foggy Cityscapes. As shown in Table~\ref{tab:1}, our proposed method PM-DETR significantly outperforms other cutting-edge approaches, achieving a 44.3\% score compared to the closest SOTA end-to-end model, MTTrans \cite{yu2022mttrans}, at 43.4\%. Furthermore, it reveals that PM-DETR considerably enhances Deformable DETR's cross-domain performance, achieving a 15.8\% absolute gain in mAP50 and outperforming all previous domain adaptive object detection methods. These promising results highlight the ability of our method to extract diverse domain-specific knowledge and effectively address distribution shift, leading to improved performance in unsupervised domain adaptation for object detection tasks. 

\noindent\textbf{Scene Adaptation.} In real-world applications, such as autonomous driving, scene layouts are not static and frequently change. It makes model performance under scene adaptation crucial. Our proposed method, PM-DETR, demonstrates its effectiveness in scene adaptation as shown in Table~\ref{tab:3}, achieving SOTA results (33.3\%) and significantly improving upon previous works. Additionally, the performance of five out of seven categories in the target domain dataset has been enhanced. 
 
\noindent\textbf{Synthetic to Real Adaptation.} The training process of object detectors using affordable and accurate simulation datasets has been proven to yield improved performance. However, this approach also brings about a notable challenge in the form of a significant inter-domain gap. In the synthetic to real adaptation scenario, we evaluated the performance of our proposed method, PM-DETR, as shown in Table~\ref{tab:2}. PM-DETR achieved state-of-the-art accuracy with a mAP of 58.6\%, outperforming Deformable DETR by 11.2\% mAP. These promising results further demonstrate the importance of a long-term domain memory space for transformer detectors to effectively extract comprehensive domain knowledge in real-world unsupervised domain adaptation scenarios.

\begin{table} [t]
  \scriptsize
  \begin{center}
\caption{Performance comparison of different methods for the synthetic to real adaptation, i.e. Sim10k to Cityscapes.}
\vspace{-0.3 cm}
\setlength{\tabcolsep}{2.2mm}{
\begin{tabular}{c|c|c|c|c}
\hline
Methods  & Detector & Publication & mAP(car) & Gain \\\hline\hline
FasterRCNN~\cite{ren2015faster}(Source)     & FRCNN & NIPS2015  & 34.6 & 00.00\\
DAF~\cite{chen2018domain}    & FRCNN  & CVPR2018 & 41.9 & +07.3\\
CR-DA~\cite{xu2020exploring}         & FRCNN  &  CVPR2020 & 43.1  & +08.5\\
DivMatch~\cite{kim2019diversify}      & FRCNN  & CVPR2019 & 43.9  & +09.3\\
SWDA~\cite{saito2019strong}          & FRCNN  & CVPR2019 & 44.6  & +10.0\\
SCDA~\cite{zhu2019adapting}         & FRCNN  & CVPR2019 & 45.1 & +10.5\\
CR-SW~\cite{xu2020exploring}          & FRCNN  & CVPR2020 & 46.2 &  +11.6\\
MTOR~\cite{cai2019exploring}           & FRCNN  & CVPR2019 & 46.6 & +12.0\\
GPA~\cite{xu2020cross}            & FRCNN  & CVPR2020 & 47.6 & +13.0\\
ViSGA~\cite{rezaeianaran2021seeking}       & FRCNN   & ICCV2021 & 49.3 & +14.7 \\\hline
FCOS~\cite{tian2019fcos}(Source)           & FCOS   & ICCV2019 & 42.5 & +7.9\\
EPM~\cite{hsu2020every}           & FCOS  & ECCV2020  & 47.3  & +12.7\\\hline
DefDETR~\cite{zhu2020deformable}(Source) & DefDETR & ICLR2021 & 47.4 & +12.8\\
SFA~\cite{wang2021exploring} & DefDETR & ACMMM2021 & 52.6& +23.3\\
MTTrans~\cite{yu2022mttrans}              & DefDETR & ECCV2022 & 57.9 & +23.3\\
\rowcolor{lightgray}~PM-DETR(Ours)              & DefDETR & - & \cellcolor{gbypink}\textcolor{red}{\textbf{58.6}} & \cellcolor{gbypink}\textcolor{red}{\textbf{+24.0}}\\\hline
\end{tabular}}
\label{tab:2}
\end{center}
\vspace{-15pt}
\end{table}

\subsection{Ablation Study}
\label{sec: exp3}

\textbf{Effectiveness of each component.} To better analyze each component in our proposed PM-DETR framework, we conduct ablation studies by accruing parts of the components in PM-DETR. As presented in Table~\ref{tab:4} (PM-DETR-AS0), the teacher-student structure is a common technique in UDA \cite{cai2019exploring, yu2022mttrans}, which is used to generate pseudo labels in the target domain and has 8.5\% mAP drop compared to our method. This verifies the improvement of our method does not come from the usage of this prevalent scheme and the model still suffers from the domain shift problem due to imperfect target domain feature extraction. In PM-DETR-AS11, by introducing prompt domain memory (PDM) in the input image, we observe that the mAP increase by 7.3\%. When employing PDM in encoder token embedding (PM-DETR-AS12) and in decoder query embedding (PM-DETR-AS13), mAP improves by 6.5\% and 6.9\%, respectively. The result clearly demonstrates that the utilization of a long-term memory space enables the model to fully learn the complex data distribution and capture diverse domain-specific knowledge in multiple levels of DETR. In terms of Prompt Memory Alignment (PMA), PM-DETR-AS21 improves the mAP to 43.8\% by encoder prompt alignment, and PM-DETR-AS22 improves the mAP 43.9\% by decoder prompt alignment. The improved performance evaluates that PMA can further reduce the discrepancy between the two domains. PM-DETR shows the complete combination of all components which achieves 15.8\%  improvement in total. It proves that all components compensate each other and jointly mitigate the object detection domain shift problem in an unsupervised paradigm.

\begin{figure*}[t]
  \centering
  \includegraphics[width=0.98 \linewidth]{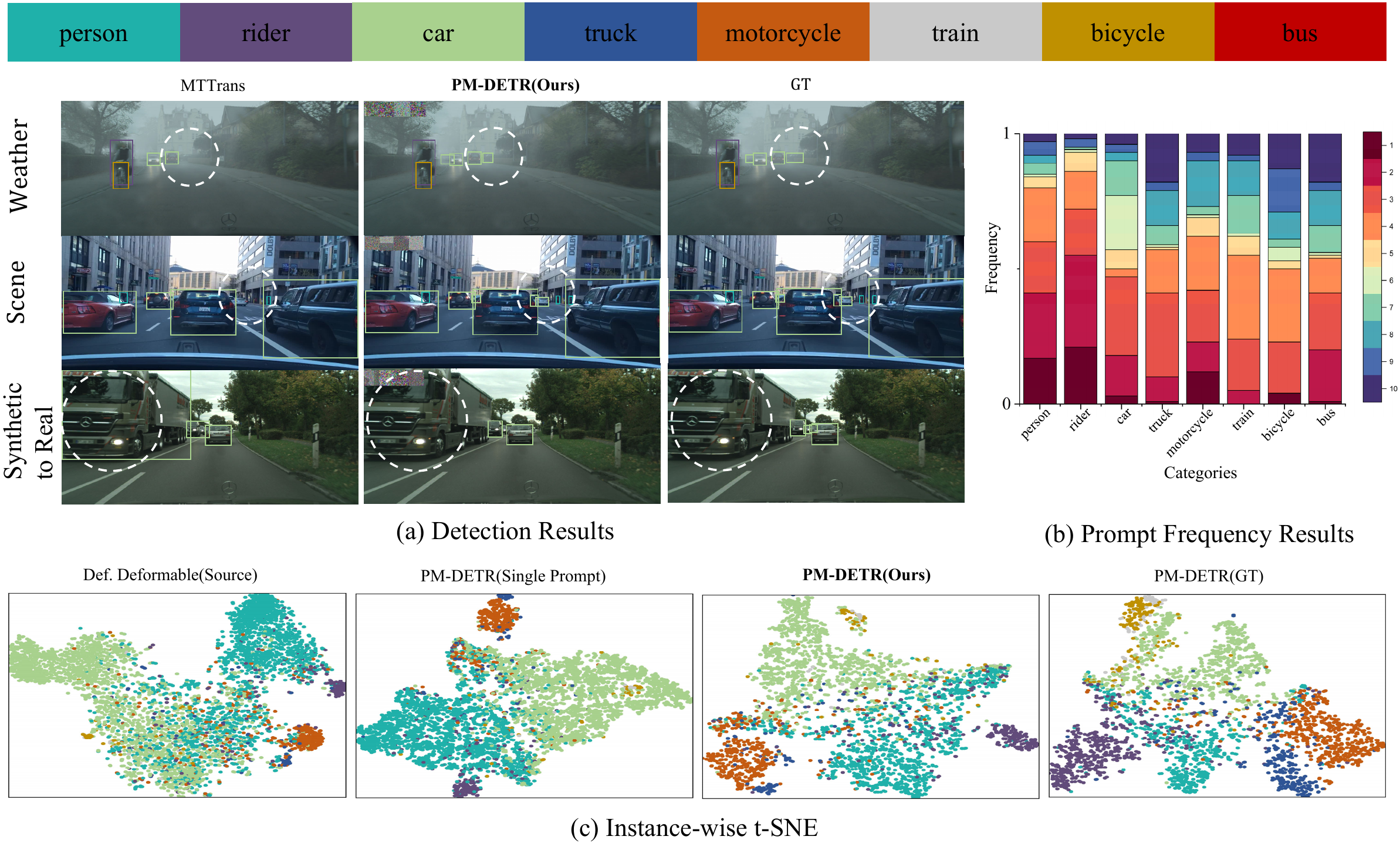}
  \vspace{-0.3 cm}
  \caption{(a) Qualitative comparison of PM-DETR with previous SOTA method and GT in three scenarios. The white circle area reflects the superiority of our method. (b) In Cityscapes to Foggy Cityscapes scenario, the frequency statistics of prompt picking in memory space correspond to different categories. (c) In Cityscapes to Foggy Cityscapes scenario, instance-level feature t-SNE results. GT stands for training in Foggy Cityscapes within fully supervised learning.}
  \label{fig:visual}
  \vspace{-0.1 cm}
\end{figure*}

\begin{figure}[htbp]
  \centering
  \includegraphics[width=\linewidth]{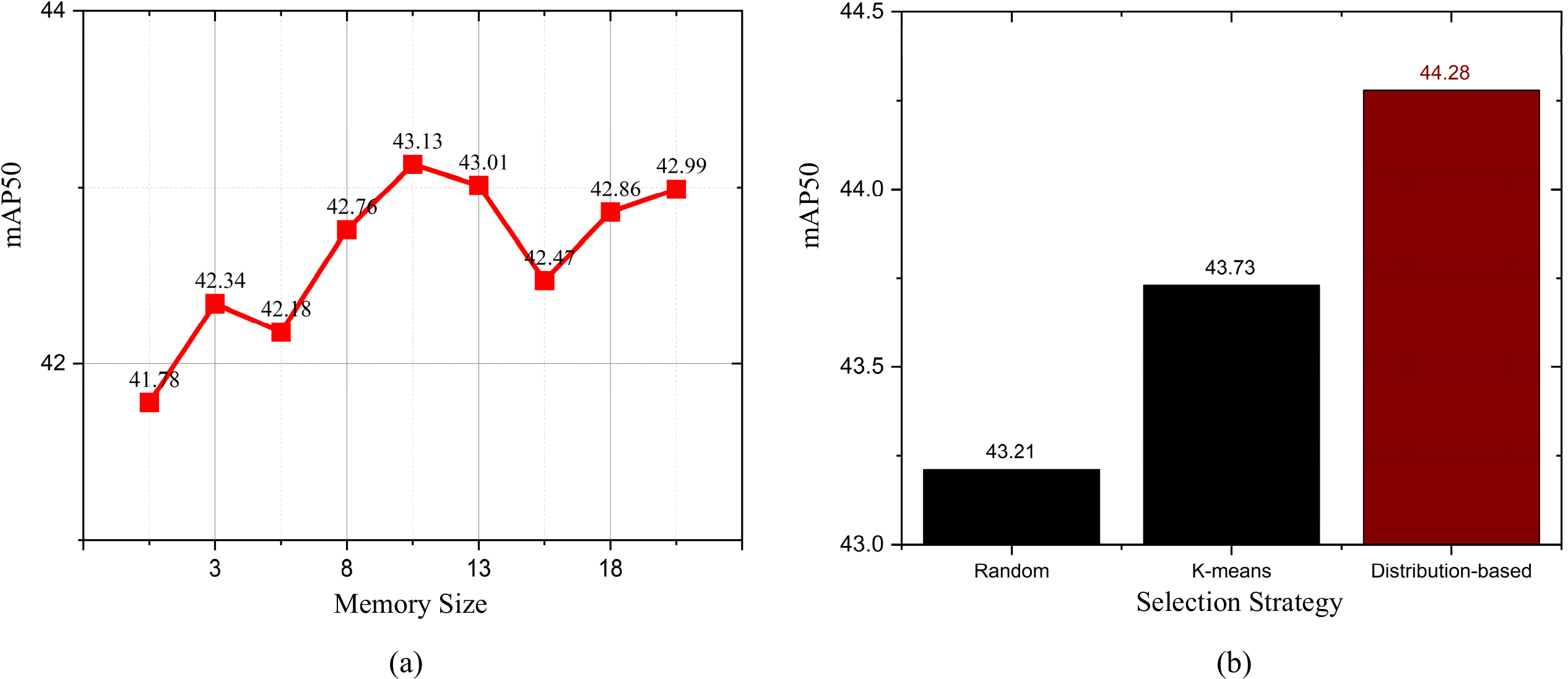}
  \vspace{-0.9cm}
  \caption{Effects of prompts’ memory size and selection strategy on the UDA task.}
  \label{fig:ablation}
  \vspace{-0.5cm}
\end{figure}

\textbf{How do the prompt memory size and selection strategy affect the performance?} In Fig.~\ref{fig:ablation} (a), we observe the impact of different prompt memory sizes on model performance. When using a single prompt (memory size equals one), there is a significant drop in performance, suggesting that a single prompt suffers severe diversity interference. As the memory size increases, the performance of the prompt-based model gradually improves and reaches its peak when the memory size is 10. Further increasing the size leads to a slight decrease in performance, but it still outperforms the single prompt scenario. This demonstrates that our prompt domain memory effectively constructs a long-term domain memory, enabling the model to understand complex data distribution and diverse domain-specific knowledge. Fig.~\ref{fig:ablation} (b) compare the effect of different prompt selection schemes on model performance, including random, k-means, and distribution-based approaches. Our distribution-based selection strategy achieves the highest performance, highlighting the effectiveness of our method in capturing crucial domain-specific information. A selection method that considers the instance-level inputs can effectively stably handle the variance of the data distribution in the target domain. To better understand the prompt selection mechanism, we plot the prompt selection frequency histograms for three domain adaptive tasks in Fig.\ref{fig:visual} (b). Our prompt selection mechanism clearly encourages more knowledge sharing between similar categories and more knowledge comparison between dissimilar categories.

\begin{table} [t]
  \centering
  \scriptsize
  \caption{Ablation studies on the weather adaptation scenario. MT stands for the mean teacher framework. Img. and Emd. stand for the input image and encoder embedding. PDM and PMA are the abbreviations of Prompt Domain Memory and Prompt Memory Alignment. Components of other experiments that differ from PM-DETR are marked in red.}
  \vspace{-0.3cm}
  \begin{tabular}{cccccccc}
  \hline
   &
     &
    \multicolumn{3}{c}{PDM} &
    \multicolumn{2}{c}{PMA} &
     \\ \cline{3-7}
  \multirow{-2}{*}{Methods} &
    \multirow{-2}{*}{MT} &
    Img. &
    Emd. &
    Query &
    $\mathcal{L}_{epa}$ &
    $\mathcal{L}_{dpa}$ &
    \multirow{-2}{*}{mAP50} \\ \hline
  Def. DETR (Source) &
    {\color[HTML]{FF0000} \XSolidBrush} &
    {\color[HTML]{FF0000} \XSolidBrush} &
    {\color[HTML]{FF0000} \XSolidBrush} &
    {\color[HTML]{FF0000} \XSolidBrush} &
    {\color[HTML]{FF0000} \XSolidBrush} &
    {\color[HTML]{FF0000} \XSolidBrush} &
    28.500 \\
  PM DETR-AS0 &
    \Checkmark &
    {\color[HTML]{FF0000} \XSolidBrush} &
    {\color[HTML]{FF0000} \XSolidBrush} &
    {\color[HTML]{FF0000} \XSolidBrush} &
    {\color[HTML]{FF0000} \XSolidBrush} &
    {\color[HTML]{FF0000} \XSolidBrush} &
    35.843 \\ \hline
  PM DETR-AS11 &
    \Checkmark &
    \Checkmark &
    {\color[HTML]{FF0000} \XSolidBrush} &
    {\color[HTML]{FF0000} \XSolidBrush} &
    {\color[HTML]{FF0000} \XSolidBrush} &
    {\color[HTML]{FF0000} \XSolidBrush} &
    43.131 \\
  PM DETR-AS12 &
    \Checkmark &
    {\color[HTML]{FF0000} \XSolidBrush} &
    \Checkmark &
    {\color[HTML]{FF0000} \XSolidBrush} &
    {\color[HTML]{FF0000} \XSolidBrush} &
    {\color[HTML]{FF0000} \XSolidBrush} &
    42.417 \\
  PM DETR-AS13 &
    \Checkmark &
    {\color[HTML]{FF0000} \XSolidBrush} &
    {\color[HTML]{FF0000} \XSolidBrush} &
    \Checkmark &
    {\color[HTML]{FF0000} \XSolidBrush} &
    {\color[HTML]{FF0000} \XSolidBrush} &
    42.765 \\
  PM DETR-AS21 &
    \Checkmark &
    \Checkmark &
    \Checkmark &
    \Checkmark &
    \Checkmark &
    {\color[HTML]{FF0000} \XSolidBrush} &
    43.812 \\
  PM DETR-AS22 &
    \Checkmark &
    \Checkmark &
    \Checkmark &
    \Checkmark &
    {\color[HTML]{FF0000} \XSolidBrush} &
    \Checkmark &
    43.943 \\\hline
  PM DETR &
    \Checkmark &
    \Checkmark &
    \Checkmark &
    \Checkmark &
    \Checkmark &
    \Checkmark &
    44.288 \\ \hline
  \end{tabular}
  \label{tab:4}
  \vspace{-5pt}
\end{table}

\subsection{Visualization and Analysis}
\label{sec: exp4}

\noindent\textbf{Detection Results.} We show some visualization results of PM-DETR on three target domain datasets, i.e. Foggy Cityscapes, BDD100k, and Cityscapes, accompanied by ground truth and previous state-of-the-art (SOTA) methods. As shown in Fig.~\ref{fig:visual} (a) Row1 (Cityscapes to Foggy Cityscapes), PM-DETR has higher recall and more accurate classification results in dense fog occlusion. As shown in Fig.~\ref{fig:visual} (a) Row2 (Cityscapes to BDD100k), our method properly classifies and locates objects even when they are heavily occluded or challengingly small in size. In Fig.~\ref{fig:visual} (a) Row3 (Sim10k to Cityscapes), we can even alleviate label misalignment (car \& truck) without supervision to some degree. All visual results are consistent with the numerical assessment results in three target domains, indicating that PM-DETR manages to mitigate the domain shift problem in the UDA Transformer detector. 



\noindent\textbf{t-SNE Distribution Results.}
Following the t-distributed stochastic neighbor embedding (t-SNE) method \cite{van2008visualizing}, in Fig.~\ref{fig:teaser} (a) and Fig.~\ref{fig:visual} (c), we visualize two types of t-SNE plots to illustrate the effectiveness of our approach: global-wise t-SNE and instance-wise t-SNE. By examining the global-wise t-SNE plot, we can gain insights into how well our method mixtures different domains. On the other hand, in the instance-wise t-SNE, we focus on visualizing individual instances and their embeddings. It can be observed that the t-SNE of our method is most similar to the results of fully supervised training in terms of inter-category distance as well as similar category aggregation. This firmly corroborates the ability of our method in mining diverse domain-specific knowledge for each category.


\section{Conclusions}

This paper presents a novel prompt-based method to enhance the adaptation ability of transformer detection by decoupling hypothesis space and mitigating the existing compromise error. Our approach leverages a hierarchical Prompt Domain Memory (PDM) to maintain a long-term memory space that facilitates comprehensive learning of the complex data distribution and diverse domain-specific knowledge. To effectively utilize PDM in cross-domain learning, we propose the Prompt Memory Alignment (PMA) method, which reduces the distribution distance between two domains by boundedly extracting the domain-specific knowledge from the memory space. We evaluate the effectiveness of our method through extensive experiments on three challenging Unsupervised Domain Adaptation (UDA) scenarios. The results demonstrate the significant improvements achieved by our approaches, PDM and PMA jointly address the distribution shift problem.
Moreover, our method is applicable across different domain distances, making it a versatile solution for various domain adaptation problems.

\newpage
\bibliographystyle{ACM-Reference-Format}
\bibliography{PM-DETR}










\end{document}